\documentclass[letterpaper, 10 pt, conference]{ieeeconf}  

\IEEEoverridecommandlockouts                              

\usepackage{graphicx} 
\usepackage{amsmath} 
\usepackage{amssymb}  

\title{\LARGE \bf
Imitation Learning for Active Neck Motion Enabling Robot Manipulation beyond the Field of View
}

\author{Koki Nakagawa$^{1}$, Yoshiyuki Ohmura$^{1}$, Yasuo Kuniyoshi$^{1}$
\thanks{$^{1}$Laboratory for Intelligent Systems and Informatics, Graduate School of Information Science and Technology, The University of Tokyo, 7-3-1 Hongo, Bunkyo-ku, Tokyo, Japan (e-mail: {k-nakagawa, ohmura, kuniyosh}@isi.imi.i.u-tokyo.ac.jp, Fax: +81-3-5841-6314)}%
}

\begin{document}

\maketitle
\thispagestyle{empty}
\pagestyle{empty}

\begin{abstract}

Most prior research in deep imitation learning has predominantly utilized fixed cameras for image input, which constrains task performance to the predefined  field of view.
However, enabling a robot to actively maneuver its neck can significantly expand the scope of imitation learning to encompass a wider variety of tasks and expressive actions such as neck gestures.
To facilitate imitation learning in robots capable of neck movement while simultaneously performing object manipulation, we propose a teaching system that systematically collects datasets incorporating neck movements while minimizing discomfort caused by dynamic viewpoints during teleoperation.
In addition, we present a novel network model for learning manipulation tasks including active neck motion.
Experimental results showed that our model can achieve a high success rate of around 90\%, regardless of the distraction from the viewpoint variations by active neck motion.
Moreover, the proposed model proved particularly effective in challenging scenarios, such as when objects were situated at the periphery or beyond the standard field of view, where traditional models struggled.
The proposed approach contributes to the efficiency of dataset collection and extends the applicability of imitation learning to more complex and dynamic scenarios.

\end{abstract}

\section{INTRODUCTION}

Deep imitation learning is an innovative method for training robots to develop object  manipulation skills based on demonstration datasets generated by human operators.
This approach proves to be particularly effective for real-world robotic applications, as it eliminates the need for manual environmental modeling or the engineering of specific behaviors and tasks.

The development of robust object manipulation strategies in unfamiliar environments and with novel objects represents a pivotal area of research.
The visual attention mechanism introduced in \cite{Kim2020} effectively predicts the direction of human gaze during task execution.
By focusing solely on the central vision image surrounding the predicted gaze location, this approach enables the exclusion of extraneous information, thereby streamlining the input for predicting robotic actions.

One challenge in transferring these object manipulation techniques to real-world applications is the reliance on fixed-viewpoint images for formulating action policies in most prior studies.
Therefore, manipulation tasks have been restricted to the perspective afforded by a stationary camera.
This constraint poses formidable challenges for object manipulation when items are located at the periphery or outside the predefined field of view.
Moreover, the effectiveness of object manipulation that leverages gaze information is contingent upon the object being positioned near the center of the visual field.
When the gaze is directed toward the edges, the accuracy of gaze measurement declines significantly, which, in turn, negatively impacts the overall performance of object manipulation tasks.

When humans perform object manipulation, their necks are not fixed, rather they can move their necks to observe the environment or notice objects beyond their original view.
The application of this approach, which enables active neck movement during object manipulation, to robotic systems effectively addresses the challenges previously identified while also broadening the scope of imitation learning to encompass a wider range of tasks.
In addition to the benefits linked to object manipulation, there are specific contexts in which a robot's neck movements carry inherent significance that must be acknowledged.
For example, intentions can be articulated through gestures using head movement \cite{Fink2012}.
Robots that are capable of dynamically adjusting their gaze and neck in response to environmental visual cues demonstrate a capacity for more naturalistic interactions with humans \cite{Rojas2021}.
In conclusion, neck motion plays a crucial role in enhancing object manipulation capabilities and enabling the execution of movements that  appear more organic.

Nevertheless, the changes in viewpoint that result from head movements may lead to distortions in visual input and adversely affect the performance of object manipulation tasks.
To date, there has been no research investigating how these changes in viewpoint influence imitation learning specifically related to object manipulation, nor their impact on the success rates of such tasks.

The contributions of this study are delineated as follows:
\begin{enumerate}
    \item We introduce a comprehensive platform for the collection of training data and present a tailored learning model dedicated for this purpose.
     \item We experimentally demonstrate that imitation learning of object manipulations that include active neck motion can achieve task success rates comparable to those achieved using conventional methods that do not include neck movements.
    \item We experimentally demonstrate that the motion of the neck significantly enhances object manipulation performance by improving the accuracy of acquiring and predicting gaze information, particularly at the edges of the visual field and beyond the camera's view.
\end{enumerate}

\section{RELATED WORK}

\subsection{Imitation Learning for Robot Manipulation}

Deep imitation learning is a method that uses a dataset that includes both environmental input data and the robot's state, paired with action output data.
This approach enables the robot to learn through supervised learning using a deep neural network.
Among the various methods of imitation learning, behavior cloning has been particularly effective; it has been successfully applied to real-world object manipulation tasks, such as grasping and lifting \cite{Zhang2018} and cloth folding \cite{Yang2017}.

The visual attention mechanism, as proposed by \cite{Kim2020}, draws inspiration from the human visual system.
During tasks, such as object manipulation, humans consistently direct their gaze toward the target object, with foveated vision providing critical detail that enhances task performance.
This eliminates the need to process all high-resolution information in the entire field of view and improves the efficiency of information processing by reducing unnecessary information.
The visual attention mechanism is an application of this framework in the domain of robotics.
In this context, only the images that surround the predicted gaze position are extracted and utilized, effectively discarding extraneous surrounding images.
This refined approach promotes the generation of robust actions in unfamiliar environments and with novel objects.
Previous research \cite{Kim2020,Kim2024da,Kim2024} has demonstrated that various object manipulation tasks can be successfully executed using gaze information.
However, none of these studies have explored scenarios where the gaze position is at the periphery of the visual field or where the camera itself is in motion.
Consequently, the impact on task performance under such conditions remains unaddressed.

\subsection{Movement of Viewpoint in Imitation Learning of Manipulation}

To the best of our knowledge, there is a paucity of research focused on integrating the dynamics of the robot's viewpoint into the imitation learning framework for object manipulation and the subsequent effects on task efficacy.

A noteworthy contribution in this area is presented by \cite{Cheng2024}, who incorporated the orientation of the robot's head during object manipulation into the action output of imitation learning.
This innovation enables the robot to execute tasks, such as picking and placing objects, across a broader spatial area than what is confined within a static camera view.
However, the robot in this study apparently did not learn to look at different directions each time depending on the object's position, and the generalization of the manipulation skill for a wide range of different object positions was not quantitatively evaluated.
Moreover, the performance of the robot was not evaluated compared to the scenarios where the head orientation was not added to the output.

\cite{Seo2023} utilized deep imitation learning to enable not only object manipulation on a desk but also locomotion with a humanoid robot.
Although their research illustrated the feasibility of imitation learning despite variations in the robot's viewpoint due to locomotion, the tasks investigated were conducted separately, without simultaneous locomotion and manipulation.

\subsection{Remote Control of Robot with Camera Movement}

The use of a head-mounted display for the remote control of a robot has proven to be an effective method for generating datasets for imitation learning \cite{Zhang2018}.
In this setup, the operator is immersed in the environment through a camera feed from the robot, projected onto the head-mounted display, while controlling the robot via a controller or an alternative device.

The operator's head angle plays a crucial role in the immersive remote control of the robot, as it enhances situational awareness through dynamic camera movement, rather than relying on a static camera feed \cite{Almeida2017}.

Nonetheless, in the field of imitation learning, there has been limited exploration of the operator's head movements.
Most studies favor images from a fixed perspective for constructing datasets intended for teaching manipulation tasks.
This oversight indicates a gap in the current research that may warrant further investigation.

\section{METHOD}

\subsection{Robot Framework}

In this study, we employed the robot teleoperation framework used in our previous studies \cite{Kim2020,Kim2024}. The framework is equipped with two UR5 robot arms to perform object manipulation tasks, although this study focused exclusively on the right arm.  The arm was outfitted with a single-degree-of-freedom (1-DOF) gripper as its end effector, enabling it to open and close for effective object grasping.

A stereo camera (ZED Mini) was positioned to align with the operator's head.
The neck mechanism of the robot featured two degrees of freedom, enabling a yaw rotation of up to 1.2 radians in both directions and a downward pitch of up to 1.2 radians.
Distinct from prior research \cite{Kim2020, Kim2024}, in which the robot's camera remained fixed,
this study incorporated a dynamic camera angle in response to the neck movements of the operator during remote control.
During testing phases this camera motion was governed by imitation learning and operated autonomously, like the robotic arm.

The human operator interacted with the robot using a controller designed to mirror the structure of the UR5.
As the operator maneuvered the controller's arm, the UR5 robot was programmed to replicate the corresponding joint angles, facilitating synchronized movements.

Additionally, the operator wore a head-mounted display (HTC Vive Pro Eye) equipped with an eye tracker to accurately monitor and record the operator's gaze direction.
The gaze data were  subsequently included in the dataset, as referenced in \cite{Kim2020}.It also features a motion-tracking system for measuring the orientation of the operator's head, which is used to control the robot's camera angle during remote control and recorded in the dataset for imitation learning as well.
Owing to the robot's neck having fewer DOF compared to that of a human neck, movements concerning neck positioning and roll are excluded from consideration, and commands are restricted to pitch and yaw angle.

\subsection{Image Projection on Head-mounted Display}

The head-mounted display
 delivers real-time images captured by the robot's stereo camera during remote operation, enabling the operator to engage with the UR5 robot as if they were physically situated within the robot's operational space.
In conventional approaches, images transmitted from the robot's camera are directly projected onto the head-mounted display in front of the operator, without accounting for the orientation of the operator's head. 

However, the present study introduces a more dynamic framework, where the orientation of the robot's camera moves according to that of the operator's head, but with a slight time lag.
Furthermore, the robot's camera is limited to 2-DoFs, so it cannot face the same direction as the head of the operator.
Therefore, with traditional methods for camera image projection, misalignment often occurs between the angles of the operator's head and their field of view, which can lead to discomfort or even nausea.

\begin{figure}[thpb]
  \centering
  \includegraphics[width=7cm]{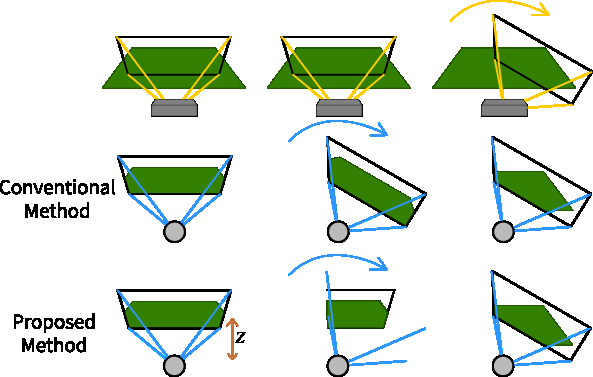}
\caption{Image projection methods in head-mounted displays.
The leftmost figure is the initial state; the center is when the operator moves his head to the right; and at the rightmost figure, the camera moves to the right after a delay.
With the proposed method, the environment can be recognized without discomfort when the angle of the operator's head is not aligned with the camera angle.
  }
    \label{decoupled}
\end{figure}

To mitigate this issue, we implement the decoupled image-projection technique as outlined by \cite{Cash2019}.
This innovative approach involves projecting the image onto a distinct virtual plane within a virtual environment, which is dynamically aligned with the current orientation of the remote camera (Fig.
\ref{decoupled}).
This setup enables the head-mounted display to deliver a stable visual perspective, effectively rectifying any misalignment between the robot's camera orientation and the operator's head orientation.
Moreover, the visual perspective can be recalibrated in real-time to reflect the angle of the operator's head at a frequency that exceeds that of the camera image refresh rate.
In contrast to the conventional method, which operates at a screen update rate of 30 Hz, corresponding to the image feed from the ZED Mini, our approach enables a screen refresh rate of 90 Hz, synchronizing with the head-mounted display's update cycle.
To the best of our knowledge, this advanced technology has yet to be employed in the generation of training data for imitation learning applications.

The ZED Mini captures an image featuring a horizontal field of view (FOV) of 108$^{\circ}$ and a vertical FOV of 57$^{\circ}$, as detailed in the accompanying datasheet.
To accurately replicate this field of view, assuming that $Z \mathrm{\,[m]}$ is the distance from the operator's eye position to the virtual plane on which the camera image is projected at its initial state, the camera image should be projected with a size of $Z \tan(108^{\circ}/2) \, \text{[m]} \times Z \tan(57^{\circ}/2) \, \text{[m]}$ in the virtual environment.
At the initial position where the image is projected in front of the operator, variations in the value of $Z$ should have no effect on the perceived field of view.
However, if the operator's posture changes, an excessively small or large change in $Z$ may cause significant distortions in the visual representation.
For this experiment, $Z=1\mathrm{\,[m]}$ was selected.

\subsection{Network Architecture}

\begin{figure*}[tbp]
\centering
\includegraphics[width=18cm,clip]{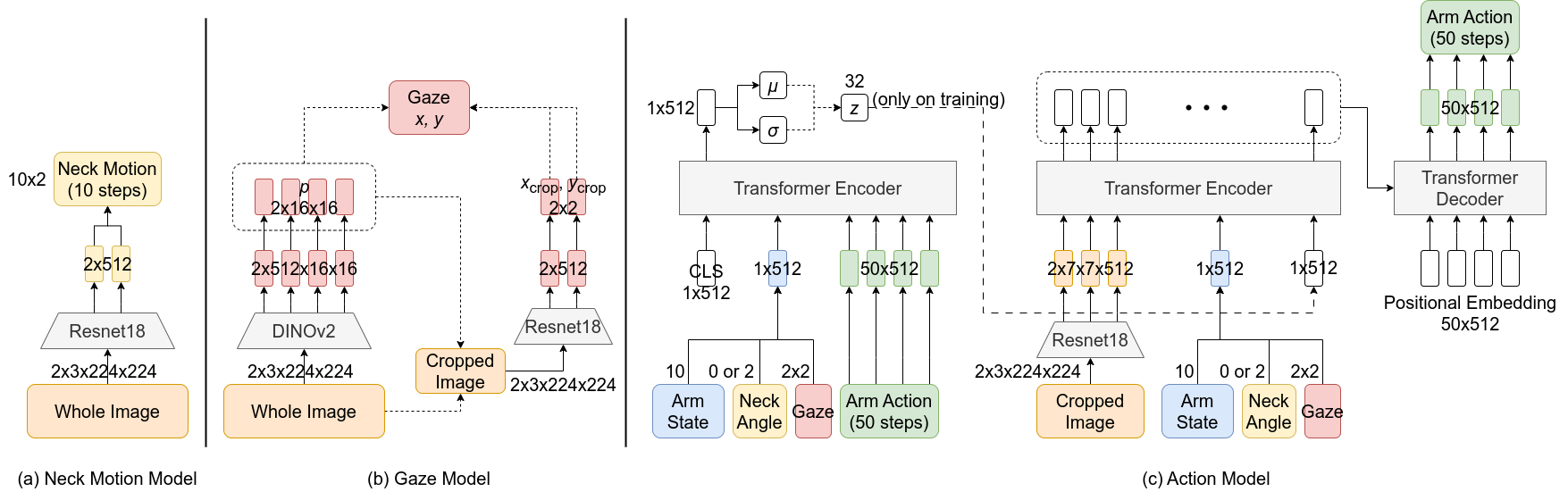}
\caption{Network model used in this study.
The model for the neck motion (a) takes an entire image as input and outputs the difference between the yaw and pitch angles of the neck.
The model for the gaze (b) comprises two sub-models: one that predicts rough locations as a classification problem (left side), and one that predicts more detailed locations based on cropped images (right side).
The model for action (c) comprises CVAE encoder and decoder as proposed in \cite{Zhao2023}.
The CVAE encoder (left side) outputs a latent variable $z$ based on the current arm state and behavior, and the CVAE decoders (center and right sides) take the image, current arm state, and latent variable $z$ as input and output the final behavior.
}
\label{fig:model_all}
\end{figure*}

The imitation learning model proposed in this research begins by predicting the neck motion and gaze direction of the operator, and subsequently infers the state of the robot's arm based on these initial predictions. The former two were predicted based on the overall image from the camera, and the latter was predicted based on the foveated image cropped with the predicted gaze position.
Therefore, the proposed model comprises three distinct sub-models, each designed to predict neck motion, gaze position, and arm state.

The neck motion prediction sub-model (Fig. \ref{fig:model_all} (a)) processes the overall image captured by the camera as input and outputs the corresponding camera angle. The input to this sub-model is features extracted from stereo images using the ResNet18 architecture \cite{he2016deep}. The output represents the differences in yaw and pitch angles of the neck between the current step and the next, effectively predicting neck motion for a duration spanning 10 steps ahead.
To determine the neck angle for the subsequent step, the Action Chunking + Temporal Ensemble method is employed. This approach calculates a weighted average of predictions for actions that occur concurrently, as detailed in \cite{Zhao2023}. 

The gaze position prediction sub-model (Fig. \ref{fig:model_all} (b)) is based on \cite{Kim2024}. It takes as input an RGB image with dimensions of $1280 \times 720$ and aims to predict gaze.
For feature extraction, DINOv2 \cite{oquab2023dinov2} is used to extract features for each of the $16 \times 16$ patches within the input image. The sub-model outputs the probability distribution of gaze presence across these patches, ultimately selecting the patch with the highest probability to determine the final gaze prediction. 
In addition, for a $224 \times 224$ region centered on the patch containing the predicted gaze position cropped from each image, another model outputs finer coordinates of the gaze.
This model utilizes features extracted by the ResNet18 architecture \cite{he2016deep} as its input, which are subsequently transformed into two outputs through a multilayer perceptron (MLP). 
The resulting predicted eye position $\hat{x},\hat{y}$ is then obtained by (1), (2); where $0 \le \hat{i}, \hat{j} \le 15$ denote the index of the patch containing the predicted gaze position, and $0 \le \hat{x_\mathrm{crop}}, \hat{y_\mathrm{crop}} \le 224$ denote the coordinates of the gaze within the cropped image.
\begin{alignat}{2}
\hat{x} &= & \, 1280 \,(\hat{i} + 0.5) / 16 +{}& (\hat{x_\mathrm{crop}} - 112)\\
\hat{y} &= & \, 720 \,(\hat{j} + 0.5) / 16 +{}& (\hat{y_\mathrm{crop}} - 112)
\end{alignat}

To predict the arm state, the model depicted in Fig. \ref{fig:model_all} (c) employs the Action Chunking with Transformer (ACT) framework\cite{Zhao2023}. 
This framework integrates a conditional variational autoencoder with a Transformer architecture, enabling precise manipulation by effectively addressing the inherent noise within the datasets. In the original research, the ACT model was designed to accept multiple images as inputs, wherein the joint angles represented the state of the robotic arm.
For our implementation, a foveated view cropped from each stereo image, aligned with the predicted gaze, was utilized as the image input. The overall state of the robot encompassed not just the status of the robotic arm, but also the current gaze coordinates and neck angles, thereby providing a comprehensive understanding of its operational context.

The state of the robot arm is represented by the position and orientation of the end effector instead of the joint angles, as it is more versatile and stable.
The representation of orientation through Euler angles, as examined in \cite{Kim2020}, introduces several complications, notably the risk of discontinuity and the occurrence of a gimbal lock depending on the configuration of the robotic arm.
To mitigate these challenges, we adopted a rotation matrix as a more robust alternative to Euler angles \cite{zhou2019}. Significantly, the third column of this rotation matrix can be uniquely derived from the other two columns, enabling the orientation to be represented by a 6-dimensional vector.
Therefore, the robotic arm's state can be encapsulated within a 10-dimensional vector, which comprises 6 components for the orientation, along with the 3D end-effector position and a single value representing the angle of the gripper.
In addition, the gaze coordinate is defined as a 4-dimensional vector, which includes the x- and y-coordinates for each eye, while the neck angle is described by the yaw and pitch angles of the neck.

The output generated from the ACT comprises a sequence that projects 50 steps ahead, formatted as 10-dimensional vectors analogous to those of  the input state.
Like the prediction of the neck motion, we used the Action Chunking + Temporal Ensemble method to determine the subsequent action.
While the original paper used the prediction results 50 steps ahead, in the experimental framework of this study we observed instability in the prediction results when forecasting 50 steps ahead. 
To address this issue, we limited the prediction scope to a weighted average of only 10 steps.

\section{EXPERIMENTS}

\subsection{Setup}

\begin{figure}[ht]
\centering
\includegraphics[width=4cm]{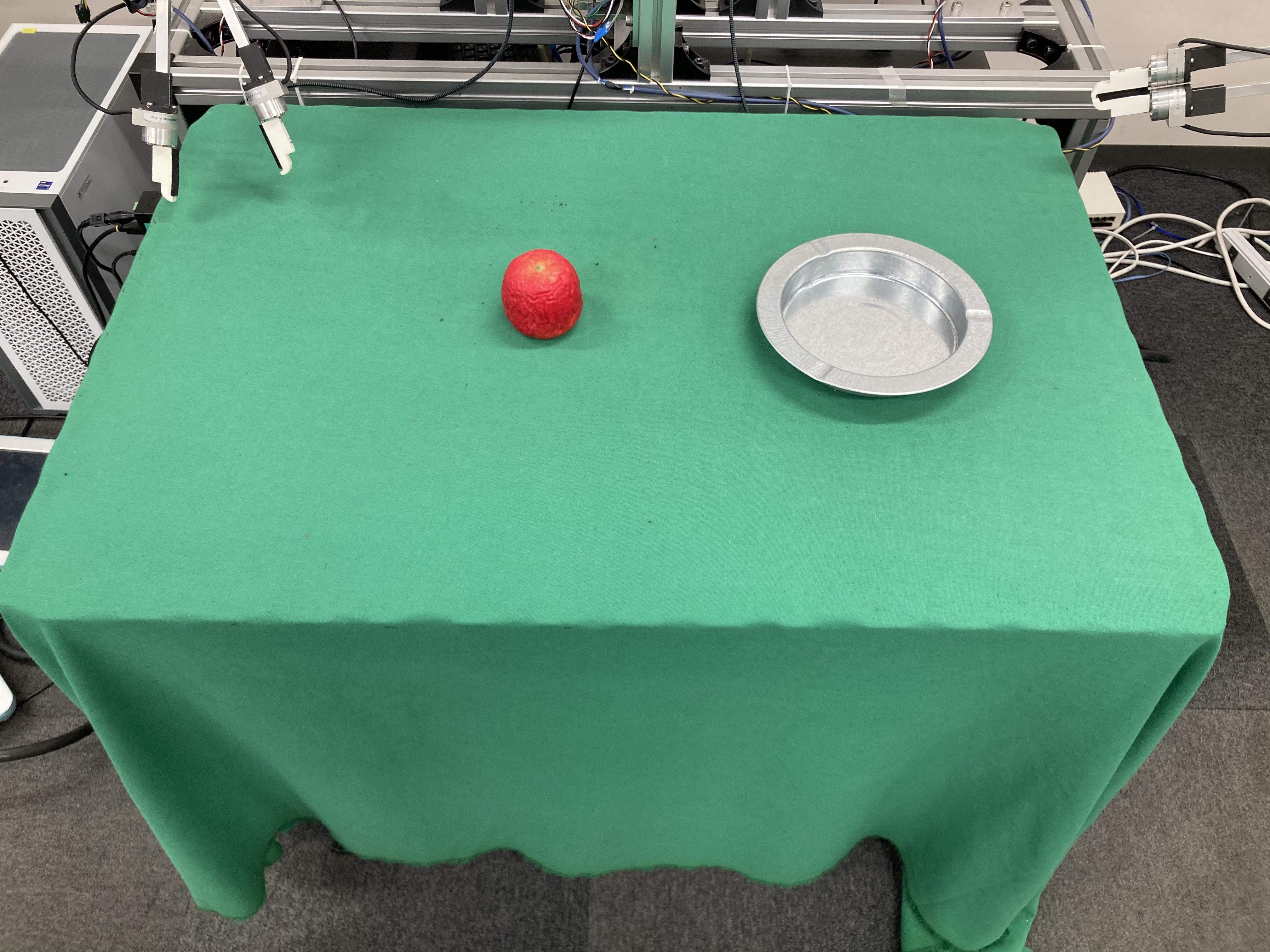}
\caption{Environment for the experiment. A desk covered with a green tablecloth, an apple-like object (left), and a silver plate (right). The image was taken from the opposite direction from the robot's camera.}
\label{fig-table}
\end{figure}

The task in this experiment was to pick up an object designed to mimic an apple, with a diameter of approximately 7 cm. 
This task was conducted on a desk measuring approximately 90 cm in width and 60 cm in depth, which was adorned with a green tablecloth (Fig. \ref{fig-table}).

The objects were strategically placed within a 45 cm $\times$ 60 cm area located on the right half of the desk as perceived by the robot. 
This arrangement compelled the robot to rotate its head in various directions based on the object position, thereby facilitating the evaluation of the robot's capacity to learn neck movements, which is the central focus of this research.

Two distinct models were utilized in the experiments: one trained on a dataset that incorporated neck movements and the other on a dataset that excluded such movements.
For the latter, we used two network sub-models from the case with neck motions described in Section 3, excluding the model that predicts neck motion and information on the neck angle from the input to the ACT.

The dataset comprised 293 episodes featuring no neck motion and 321 episodes involving neck motion.
The dataset without neck motion contained a slightly lower number of episodes compared to the dataset with neck motion, which can be attributed to the omission of instances where the object was completely out of view.
For both subsets, 90\% of the data was designated for training purposes, while the remaining 10\% was utilized for testing. 

\begin{figure}[ht]
\centering
\includegraphics[width=6cm]{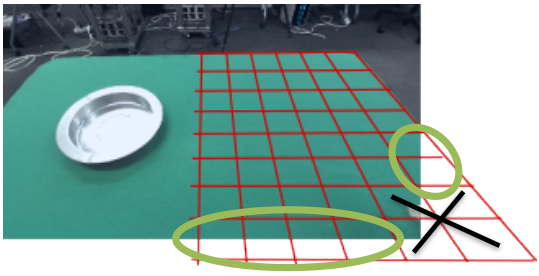}
\caption{Experimental environment as seen from the robot's side. The approximate location of the apple in the experiment is indicated by the grid. The area circled in green is defined as out of view, and no experiments were performed in the areas marked with X.}
\label{table_from_robot}
\end{figure}

During the testing phase, the model trained using two distinct random seeds for each dataset was tasked with executing the task twice for each of 44 designated locations within the right half of the desk area.
As illustrated in Fig. \ref{table_from_robot}, positions where the object overlapped with the arm's initial state and were not feasible for placement were systematically excluded from consideration.

\subsection{Result}

\begin{figure*}[ht]
\centering
\includegraphics[width=16cm]{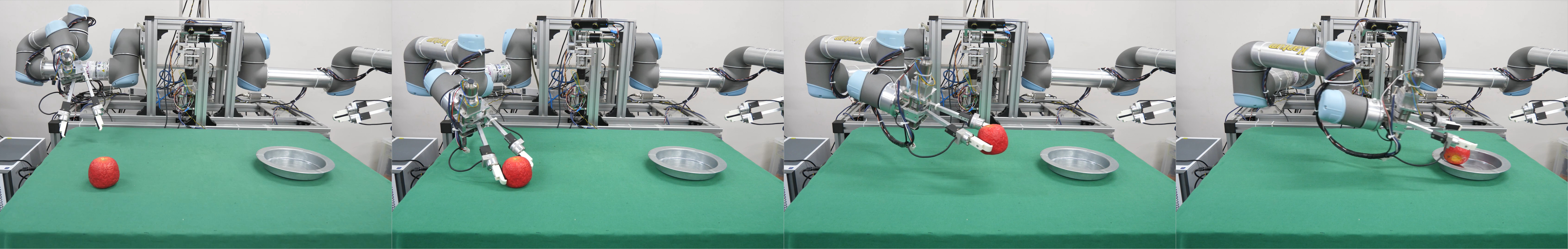}
\caption{Example of the robot performing the task with active neck motion, where the robot turns its head in the direction of the object.}
\label{pic_clip}
\end{figure*}

The proposed methodology demonstrated its capability to effectively learn how to pick up and transport objects located across a diverse range of positions, while concurrently changing the neck angle in response to the object placement (Fig. \ref{pic_clip}).

\begin{figure}[ht]
\centering
\includegraphics[width=7cm]{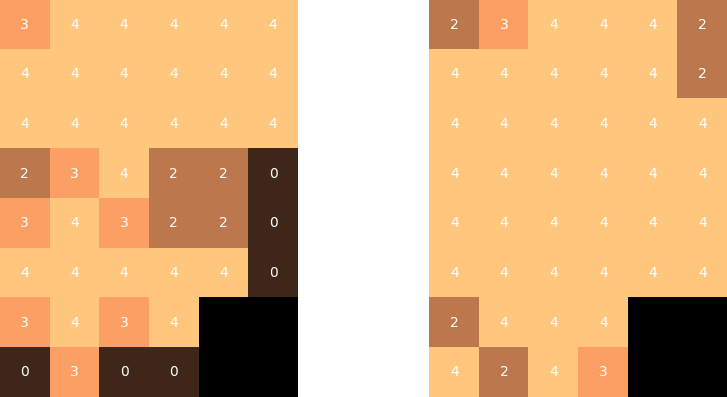}
\caption{Number of successful attempts at the task without neck motion (the left figure) and with neck motion (the right figure), out of 4 attempts for each location.}
\label{fig:heatmap}
\end{figure}

Among the 176 trials, 152 trials were conducted with the object positioned inside the view, and in the remaining 24 trials, the object was located at the edge or outside of the view. Accordingly, we tabulated the number of times the task was successfully completed (Table \ref{table:total}, Fig. \ref{fig:heatmap}).
“Out of view” refers to when the object was partially or completely outside of the view (Fig. \ref{table_from_robot}).
“In view” refers to the results of the experiment when the object is in any other location.

\begin{table}[ht]
  \caption{Total task success rate for each dataset. The numbers represent the number of successful attempts at the task out of a total of 176 attempts, 152 in view and 24 out of view, respectively.}
  \label{table:total}
  \begin{center}
    \begin{tabular}{ccccccc}
      \hline
      \multicolumn{2}{c}{In View} &
      \multicolumn{2}{c}{Out of View} &
      \multicolumn{2}{c}{Total}
      \\
      \hline
      Without Neck & 132 & (86.8\%) & 3 & (12.5\%) & 135 & (76.7\%) \\
      \textbf{With Neck} & 143 & (94.1\%) & \textbf{21} & \textbf{(87.5\%)} & 164 & (93.2\%) \\
      \hline
    \end{tabular}
  \end{center}
\end{table}

\begin{figure}[ht]
\centering
\includegraphics[width=8.5cm]{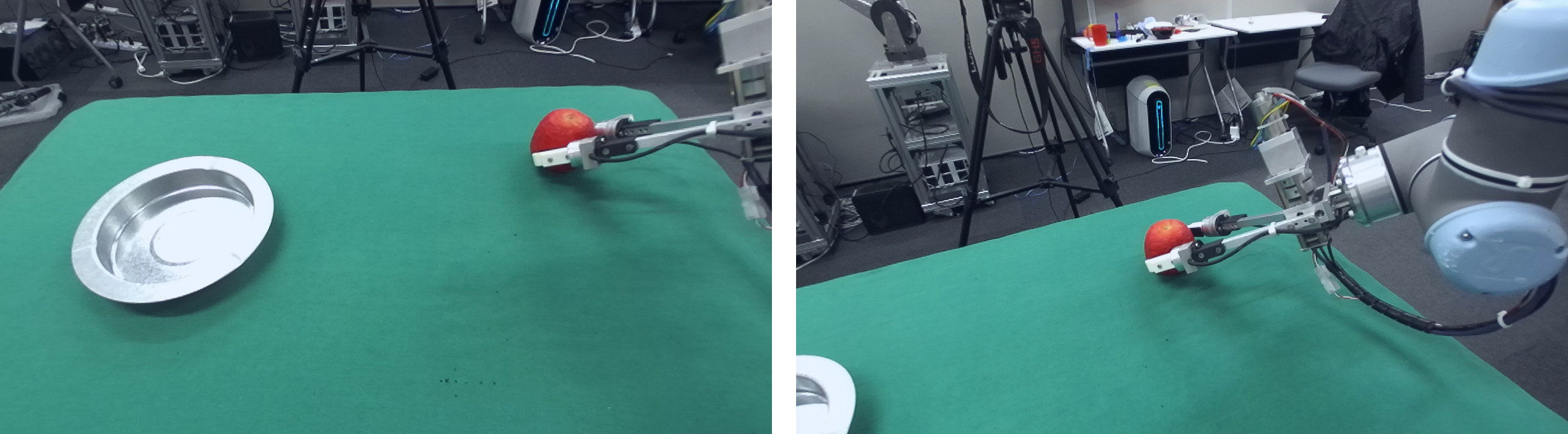}
\caption{View from the camera on the robot.
The right side image is taken during the task with neck motion, which contains many irrelevant objects from outside the desk, compared to the left side image, which shows the task without neck motion.
}
\label{disturbance}
\end{figure}

For experiments conducted with the object in view, high success rates of around 90\% are achieved, with and without neck movements.
Notably, although neck motion introduced additional disturbances from outside the desk within the field of view of the camera (Fig. \ref{disturbance}), the result showed that neck movements can be incorporated into imitation learning for object manipulation without significantly compromising task performance.

\addtolength{\textheight}{-0cm}   

\begin{figure}[t]
\centering
\includegraphics[width=8.5cm]{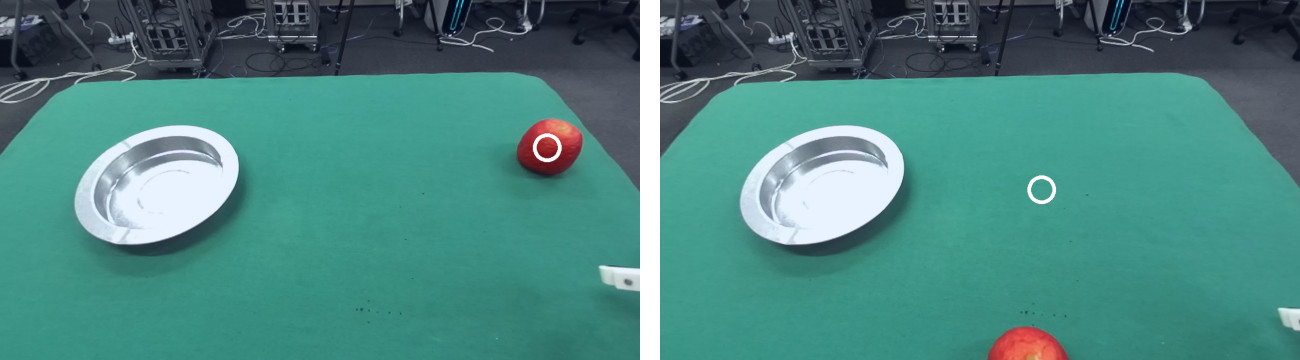}
\caption {Example of the result of gaze prediction. The white circle represents the predicted gaze position.
Normally, the position of the object and the gaze overlap exactly as shown in the left figure,
but when the object is at the edge of the view, the result of the prediction of the gaze information may be wrong as shown in the right figure.}
\label{gaze_prediction_ex}
\end{figure}

In contrast, the results of experiments involving objects outside the visible field were significantly poorer, achieving a success rate of only 12.5\% when neck motion was not utilized.
When the operator's gaze was directed toward the periphery of the view, the ability to accurately measure gaze position diminished, leading to imprecise gaze data included in the dataset (Fig. \ref{gaze_prediction_ex}). 
Consequently, if an object is situated at the edge of the field of view, the robot struggles to accurately predict the gaze position, which ultimately results in flawed actions.
However, the proposed method with neck movements achieved a high success rate comparable to that of the in-view condition (87.5\%), even on the outside of the view condition.
Active neck motion plays a crucial role in preventing gaze-prediction failures by ensuring that objects are positioned within the central visual field. This approach significantly enhances performance during object manipulation, particularly when the target object is located at the periphery or outside the visual field.

\section{DISCUSSIONS}

In this study, we proposed a novel platform designed for the collection of training datasets for manipulation task including active neck motion, and an imitation learning model that learns to move its neck motion.

The experimental findings indicate that we can achieve high success rates in object-manipulation tasks---rates that are comparable to those observed in static viewpoint scenarios---despite variations in camera angles.
Furthermore, we have demonstrated that enhancing the accuracy of gaze information prediction is achievable, leading to impressive success rates in manipulations involving objects positioned beyond the field of view.
This is particularly significant, as such scenarios are difficult to manage using traditional methods that do not account for neck movement.
Additionally, the incorporation of head movement offers distinct advantages in dataset collection, enabling human operators to execute tasks more effectively when objects are located at the edges or completely outside their line of sight. 
In summary, the integration of neck movement in robotic manipulation expands the range of tasks suitable for imitation learning, while maintaining task performance levels.

Although the simplified neck mechanism presumed in this study pose certain limitations, future research can explore more complex tasks and adaptive learning.
The object manipulation task conducted herein was performed on a desk; however, the ability to move the neck significantly expands the robot's visual field, encompassing not only the area surrounding the desk but also the space above it. 
Additionally, a camera with enhanced degrees of freedom is expected to facilitate more advanced behaviors, including the ability to switch viewpoints to aid task execution, explore the environment, and convey intentions to others. Future research efforts will focus on how to implement these behaviors through imitation learning methodologies.






\bibliographystyle{IEEEtran}
\bibliography{IEEEabrv,mybibfile}

\end{document}